\documentclass[runningheads]{llncs}
\usepackage[T1]{fontenc}
\usepackage{cite}
\usepackage{amsmath,amssymb,amsfonts}
\usepackage{algorithmic}
\usepackage{graphicx}
\usepackage{textcomp}
\usepackage{xcolor}
\usepackage{subcaption}
\usepackage[most]{tcolorbox} 
\usepackage{ragged2e}
\usepackage{lipsum}
\usepackage[font=scriptsize]{caption}
\newtcolorbox{llmbox}[1][]{
  enhanced,
  breakable,                 
  arc=4pt, 
  colback=blue!3,            
  colframe=blue!60!black,    
  coltitle=blue!20!black,
  coltext=red!50!black, 
  fontupper=\ttfamily\small, 
  fonttitle=\bfseries, 
  left=1.2ex, right=1.2ex, top=0.8ex, bottom=1ex,
  boxsep=0.6ex,
  title=LLM Output,          
  #1                         
}

\begin{document}
\title{The Agentic Leash:  Extracting Causal Feedback Fuzzy Cognitive Maps with Mixed Large Language Models}
\titlerunning{Agentic Learning of Feedback Causal FCMs with LLM Agents}
%
\author{Akash Kumar Panda \inst{1} \and
Olaoluwa Adigun \inst{2} \and
Bart Kosko \inst{3}}
\authorrunning{P. Akash et al.}
%
\institute{University of Southern California, Los Angeles, CA 90007, USA  \\ \email{akashpan@usc.edu} \and
Florida International University, Miami, FL 33199, USA \\
\email{olaadigu@fiu.edu}  \and
University of Southern California, Los Angeles, CA 90007, USA \\ \email{kosko@usc.edu}}
\maketitle              
\begin{abstract}
We design a large-language-model (LLM) agent system that extracts causal feedback fuzzy cognitive maps (FCMs) from raw text.
The causal learning or extraction process is agentic both because of the LLM's semi-autonomy and because ultimately the FCM dynamical system's equilibria drive the LLM agents to fetch and process causal text.
The fetched text can in principle modify the adaptive FCM causal structure and so modify the source of its quasi-autonomy---its equilibrium limit cycles and fixed-point attractors. 
This bidirectional process endows the evolving FCM dynamical system with a degree of autonomy while the system still stays on its agentic leash.
We show in particular that a sequence of three system-instruction sets guide an LLM agent as it systematically extracts key nouns and noun phrases from text, as it extracts FCM concept nodes from among those nouns and noun phrases, and then as it extracts or infers partial or fuzzy causal edges between those FCM nodes. 
We test this FCM generation on a recent essay about the promise of AI from the late diplomat and political theorist Henry Kissinger and his colleagues.
This three-step process produced FCM dynamical systems that converged to the same equilibrium limit cycles as did the human-generated FCMs even though the human-generated FCM differed in the number of nodes and edges. 
A final FCM mixed generated FCMs from separate Gemini and ChatGPT LLM agents.  
The mixed FCM absorbed the equilibria of its dominant mixture component but also created new equilibria of its own to better approximate the underlying causal dynamical system.
\end{abstract}

\section{The Agentic Leash:  Growing Fuzzy Cognitive Map Dynamical Systems from Text}

We show how agentic passes through structured LLM agents can grow causal feedback fuzzy cognitive maps (FCM) from sampled text documents. 

These causal FCM feedback dynamical systems form local fuzzy or partial causal rules from the sampled documents.
This local causal structure in turn defines global equilibrium limit cycles that serve as scenario-like answers to causal what-if questions. 
They also define the very source of the FCM dynamical system's agency -- its evolving equilibrium limit cycles.
This differs from ordinary feedforward agentic systems whose agency resides only in programmed commands.
Mixing FCMs can give both richer learned causal knowledge bases and richer global equilibria.
The extraction process is agentic \cite{acharya2025agentic} or partially autonomous because the FCM's evolving global equilibria command the LLM agents to fetch and process further text that then tends to change the commanding FCM equilibria.  
Related work uses an autoencoder-like mapping to convert FCMs to text and continue the reverberatory process \cite{panda2025causal}.
This bidirectional process keeps the FCM's LLM agents on a type of flexible agentic leash.

\begin{figure}[!ht]
\centering
\includegraphics[width=0.9\textwidth]{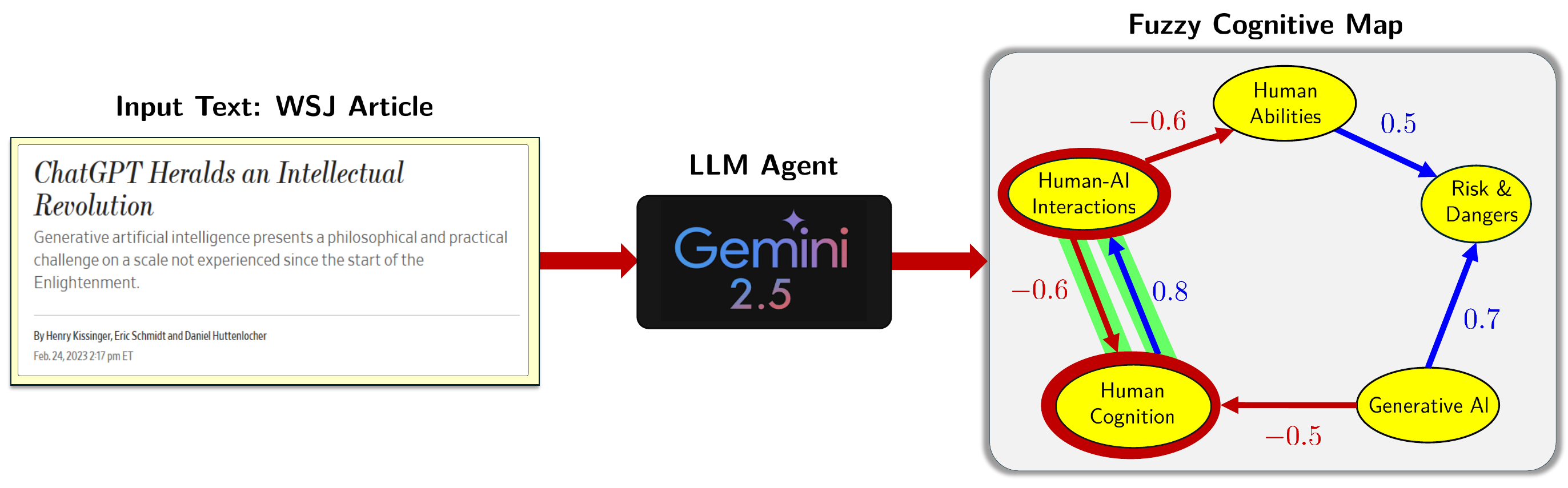}
\caption{\footnotesize{A Large Language Model (LLM) extracts causal variables and their causal relationships out of a \emph{Wall Street Journal} article from Henry Kissinger and colleagues about the promise of AI and then creates a Fuzzy Cognitive Map (FCM).
The figure shows only 5 out of the 15 AI-extracted nodes and the directed weighted edges that connect them. 
The positive edges are in blue and the negative edges are in red. 
The figure highlights one of many feedback loops in the FCM in green.
In this case:  Growth of \emph{Human Cognition} increases \emph{Human-AI Interactions} but an increase in \emph{Human-AI Interactions} decreases \emph{Human Cognition}.}
}
\label{fig:WSJ-to-FCM}
\end{figure}
\begin{figure}[ht]
\centering
\includegraphics[width=0.9\textwidth]{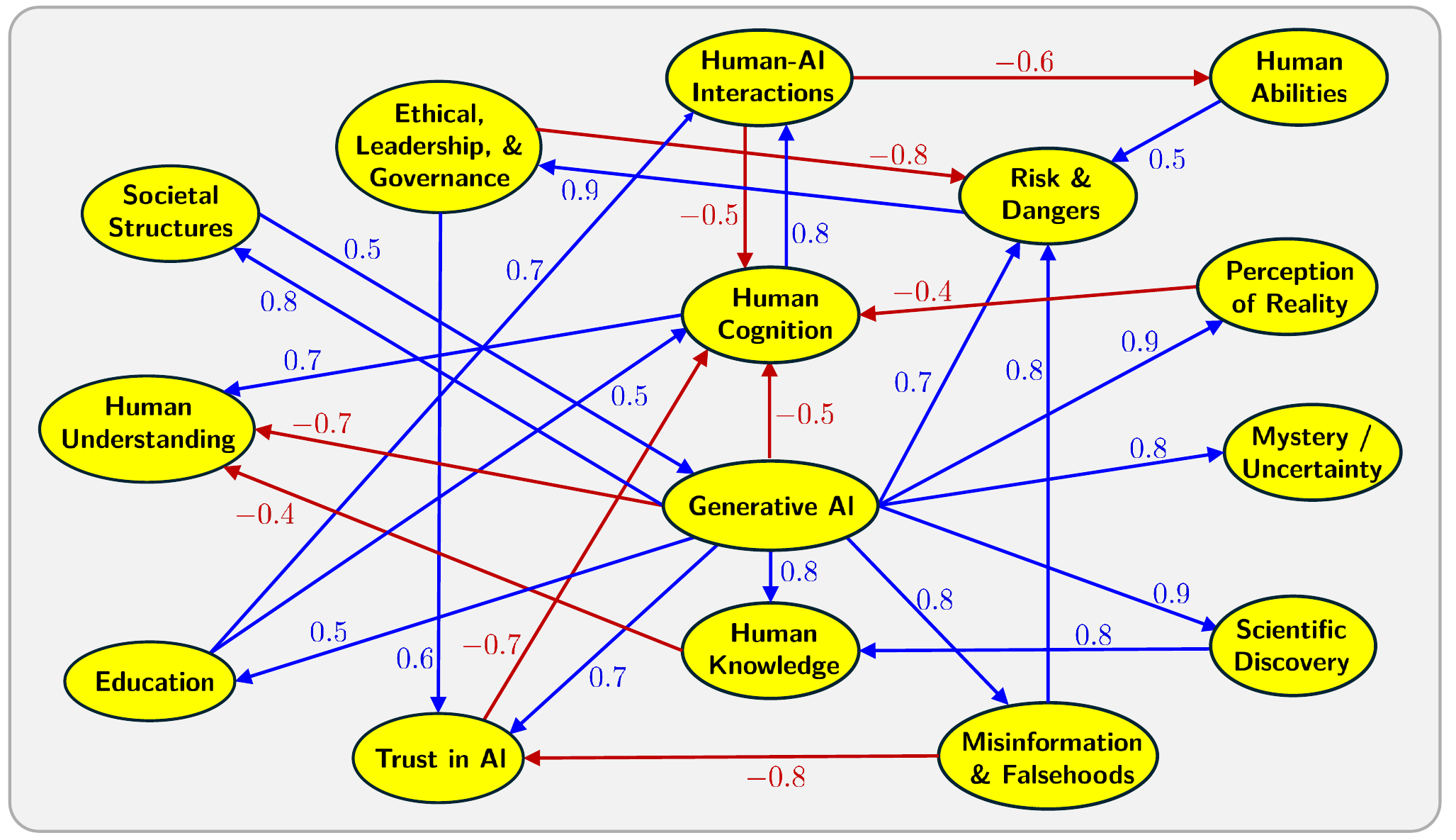}
\caption{A 15-node FCM extracted by the LLM from the WSJ article titled \emph{``ChatGPT Heralds an Intellectual Revolution''} by Henry Kissinger et al.
This FCM converges to a 4-step limit cycle. 
The limit cycle predicts that the growth of generative AI comes in waves or cycles. 
In the $1^{\text{st}}$ step generative AI gets widely used, human-AI interactions rise, and human knowledge grows. 
This helps leaders govern ethically in the $2^{\text{nd}}$ step but it also comes with risks and dangers.
In the $3^{\text{rd}}$ step: People trust the mysterious and uncertain AI, misinformation and falsehoods spread, and society changes. 
But AI improves education at the same time and leads to scientific discoveries. 
Generative AI is used widely again in the $4^{\text{th}}$ step but this time without ethical leadership. 
Society changes again but this time without generative AI. 
}
\label{fig:kissinger-FCM}
\end{figure}

\begin{figure}[ht]
\centering
\includegraphics[width=\textwidth]{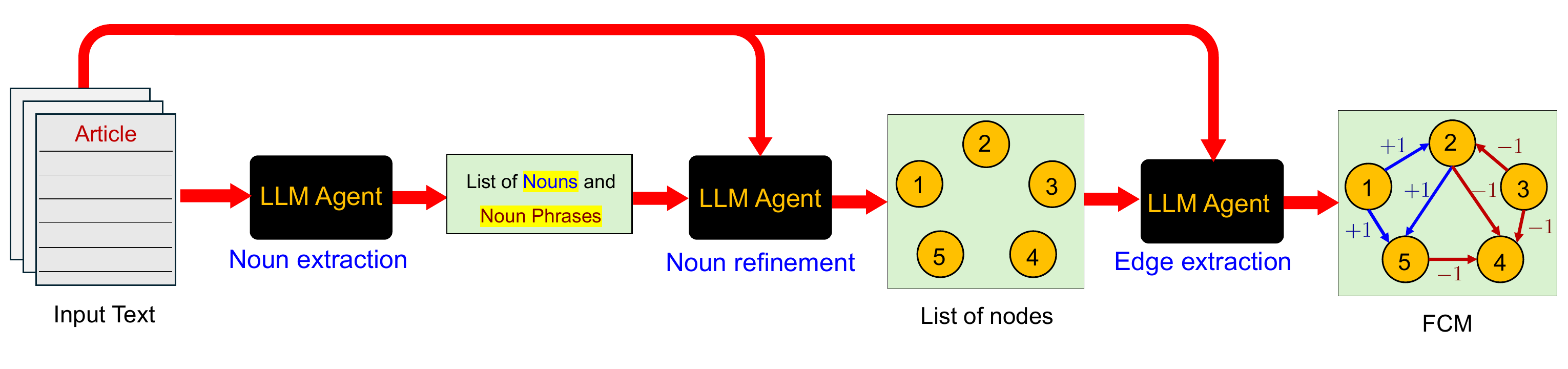}
\caption{\footnotesize{LLM-based FCM-extraction from text: 
An LLM agent  extracts FCM nodes and edges from a text input through a three-step systematic process. 
These steps use the same LLM agent with guiding system instructions. 
The LLM extracts all the nouns and noun phrases used in the text during Step 1. 
It extracts the FCM nodes from the list of nouns and noun phrases in Step 2.
This involves refining the noun list from Step 1 based on associated qualitative or quantitative measures as well as the existence of causal links. 
The LLM extracts fuzzy-weighted edges from a list of node-pairs to complete the FCM in Step 3.}}
\label{fig:LLM-FCM-EXtractor}
\end{figure}

Figure \ref{fig:WSJ-to-FCM} shows a feedback causal sub-network of the complete 15-node FCM in Figure~\ref{fig:kissinger-FCM}.
 An LLM agent grew the complete FCM from a recent AI article titled \emph{``ChatGPT Heralds an Intellectual Revolution''} by Henry Kissinger et al. in the \emph{Wall Street Journal} \cite{kissinger2023chatgpt}.  
We note that the first published FCM in the original FCM paper \cite{kosko1986fuzzy} described key variables of Mideast peace based on the author's reading of a 1982 syndicated newspaper opinion piece from Henry Kissinger.
The green highlight shows just one embedded feedback loop that traces the causal flow from \emph{Human-AI Interaction} to {\emph{Human Cognition}} and then back to itself:  \emph{Human-AI Interaction} \textcolor{red}{$\rightarrow$} \emph{Human Cognition} \textcolor{blue}{$\rightarrow$} \emph{Human-AI Interaction}. 
Here ``\textcolor{red}{$\rightarrow$}'' denotes a negative edge and ``\textcolor{blue}{$\rightarrow$}'' denotes a positive edge. 
Even this smaller 5-node sub-FCM encodes equilibrium limit cycles that can serve as answers to what-if questions.  
Figure~\ref{fig:Gemini-Limit-Cycles} below shows this process for a simpler FCM.

These learned FCM knowledge graphs consist of local causal rules.  
The rules give an immediate local form of interpretability or explainable AI (XAI)\cite{8466590,guimera2020bayesian,horst2019explaining,samek2019explainable,giamattei2025causal}.  
The embedded or hidden limit cycles in feedback FCMs also give a global form of XAI.

A key problem is that text articles usually discuss the causal variables and causal relationships of an underlying dynamical system without explicitly mentioning this system or its dynamics. 
Authors may have to guess how the feedback causal-rule based system behaves under given conditions.
Authors or speakers may not even be aware that their causal portraits define nonlinear feedback dynamical systems.
Nor does the author or the reader have an easy way to see the system's dynamics or verify the author's equilibrium predictions even if they are aware of the feedback structure.

A further problem in FCM modeling is that it can be expensive and time consuming to hire an expert to read the text and correctly predict the dynamics of the system.
A large-scale FCM may well be too complex for such predictions. 
Large Language Models (LLMs) can instead parse the body of text and then process it given prompts from users or from the evolving FCM dynamical system itself.

\begin{figure}[ht]
\centering
\includegraphics[width=0.85\textwidth]{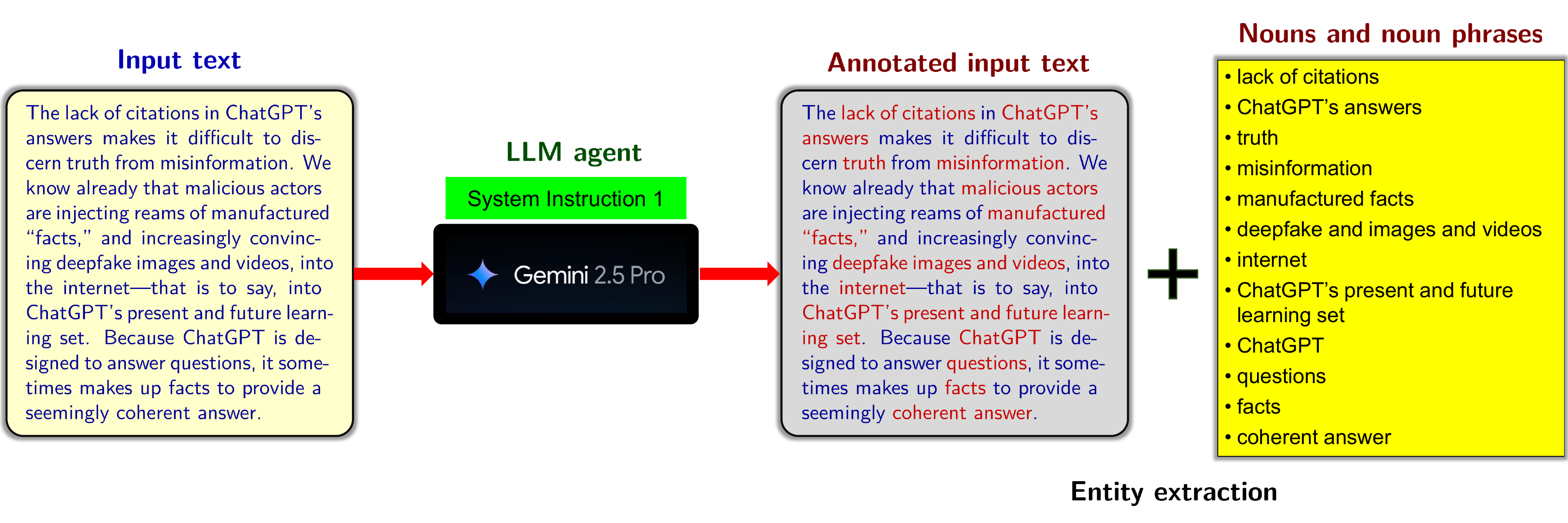} 
\caption{Noun Extraction with an agentic LLM: 
The LLM takes a paragraph of text as input and identifies the nouns, noun phrases, and pronouns present in the text. 
The figure on the right highlights the extracted nouns and noun phrases in red.}
\label{fig:LLM-Agent-1}
\end{figure}

We need an LLM that can systematically extract causal variables and their causal relationships from the text without human supervision. 
These causal rules from the text should then predict the dynamics of the system described by the text even if the text does not mention the dynamics. 
FCMs can model this causal information as a weighted directed cyclic graph and also give qualitative equilibrium predictions from the equilibria of the approximating causal FCM \cite{kosko1986differential,kosko1986fuzzy,kosko1988hidden,osoba2017fuzzy,ziv2018potential,glykas2010fuzzy,papageorgiou2013fuzzy,stach2010divide,kosko1988hidden,taber2007quantization}.

We show that LLM agents can systematically extract FCM nodes and edges from a body of text through a sequence of finely tuned system instructions. 
The LLMs can give reasons for their decisions by quoting from the text source.
These textual anchors can also help reduce hallucinations. 

\begin{figure}[ht]
\centering
\includegraphics[width=0.85\textwidth]{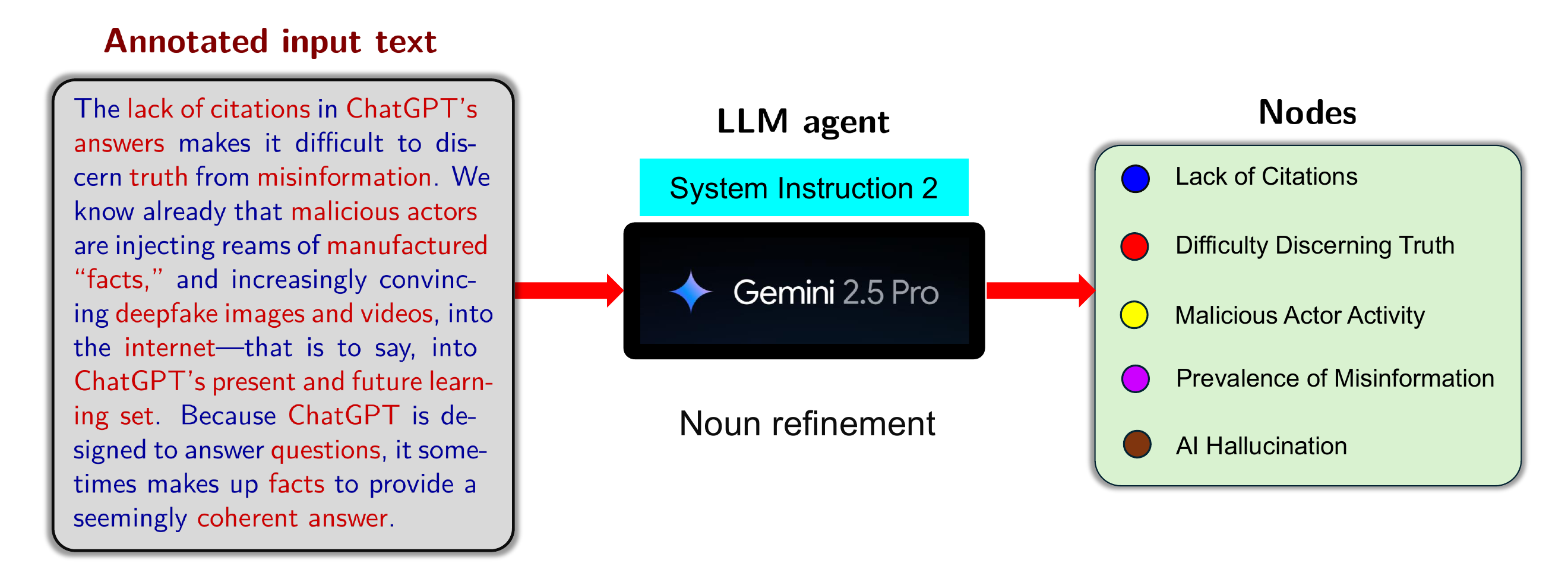}
\caption{FCM-node extraction with an agentic LLM: 
The LLM takes the list of nouns and noun phrases as input and filters those that are associated with some kind of qualitative or quantitative measure.
This gives a list of 5 FCM nodes. }
\label{fig:LLM-Agent-2}
\end{figure}

Figure~\ref{fig:WSJ-to-FCM} shows the simplest version of the FCM agentic set-up. 
An LLM takes the WSJ article \emph{``ChatGPT Heralds an Intellectual Revolution''} by Henry Kissinger \emph{et al} as text input and extracts causal variables and causal relationships among those variables from the text. 
It then builds an FCM out of those causal rules. 
Figure~\ref{fig:WSJ-to-FCM} shows only 5 of the 15 nodes from the extracted FCM. 
Figure~\ref{fig:kissinger-FCM} shows the complete FCM.

Section~\ref{FCM} explains FCMs and shows how FCMs can model or approximate causal dynamical systems. 
The section explains what the FCM nodes and edges mean and how the FCM evolves in discrete time and converges to its equilibria.  
FCMs can also evolve in continuous time.

Section~\ref{LLM-FCM-Extractor} explains the three-step systematic process to extract an FCM from a body of text that describes a dynamical system. 
It explains the 3 respective prompts to the LLM agent that detect nouns and noun phrases in the text, that extract nodes from the nouns, and that extract edges from the node-pairs. 
Figure~\ref{fig:LLM-FCM-EXtractor} shows this process.

\begin{figure}[ht]
\centering
\includegraphics[width=0.85\textwidth]{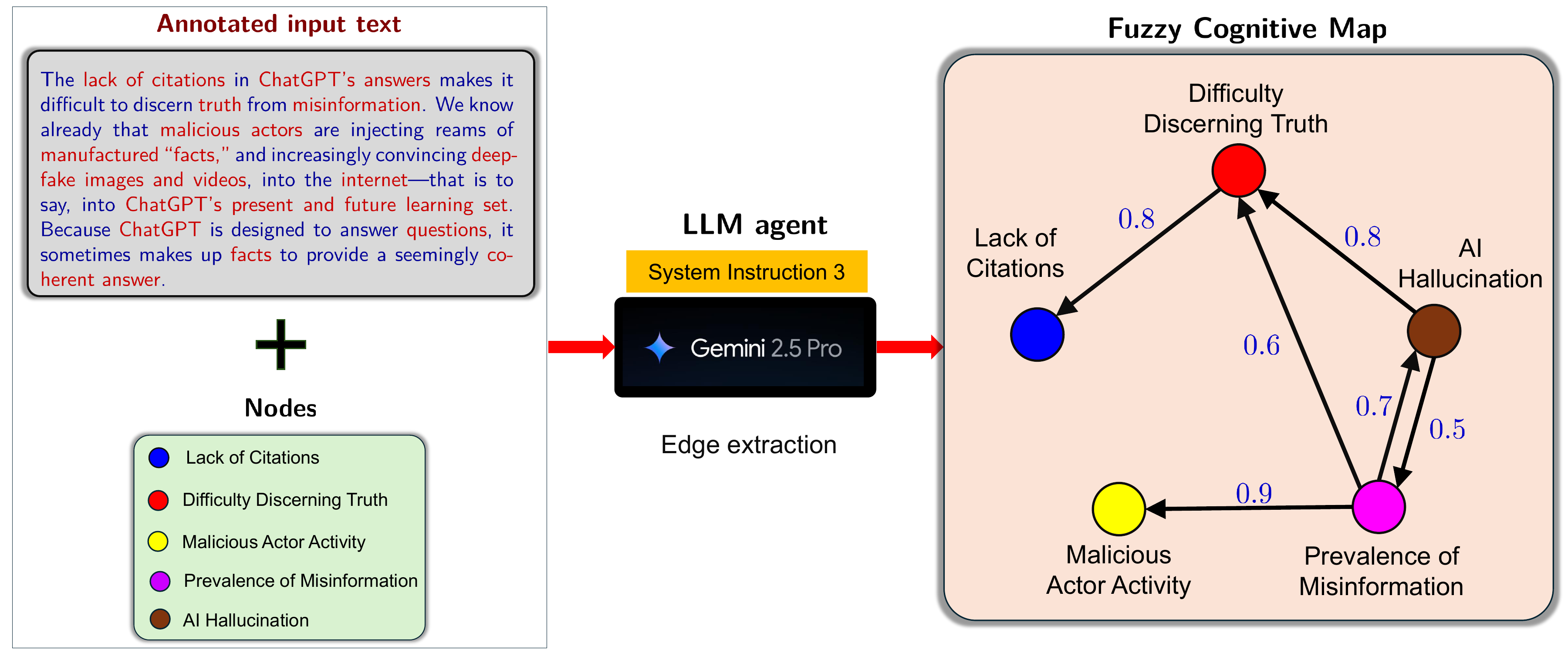}
\caption{FCM-edge extraction with an agentic LLM: 
The LLM goes through node-pairs and looks for textual evidence for causal connection. 
This gives the FCM 6 causal edges. }
\label{fig:LLM-Agent-3}
\end{figure}

\begin{figure}[!ht]
\centering
\begin{subfigure}{0.5\linewidth}
\includegraphics[width=0.96\textwidth]{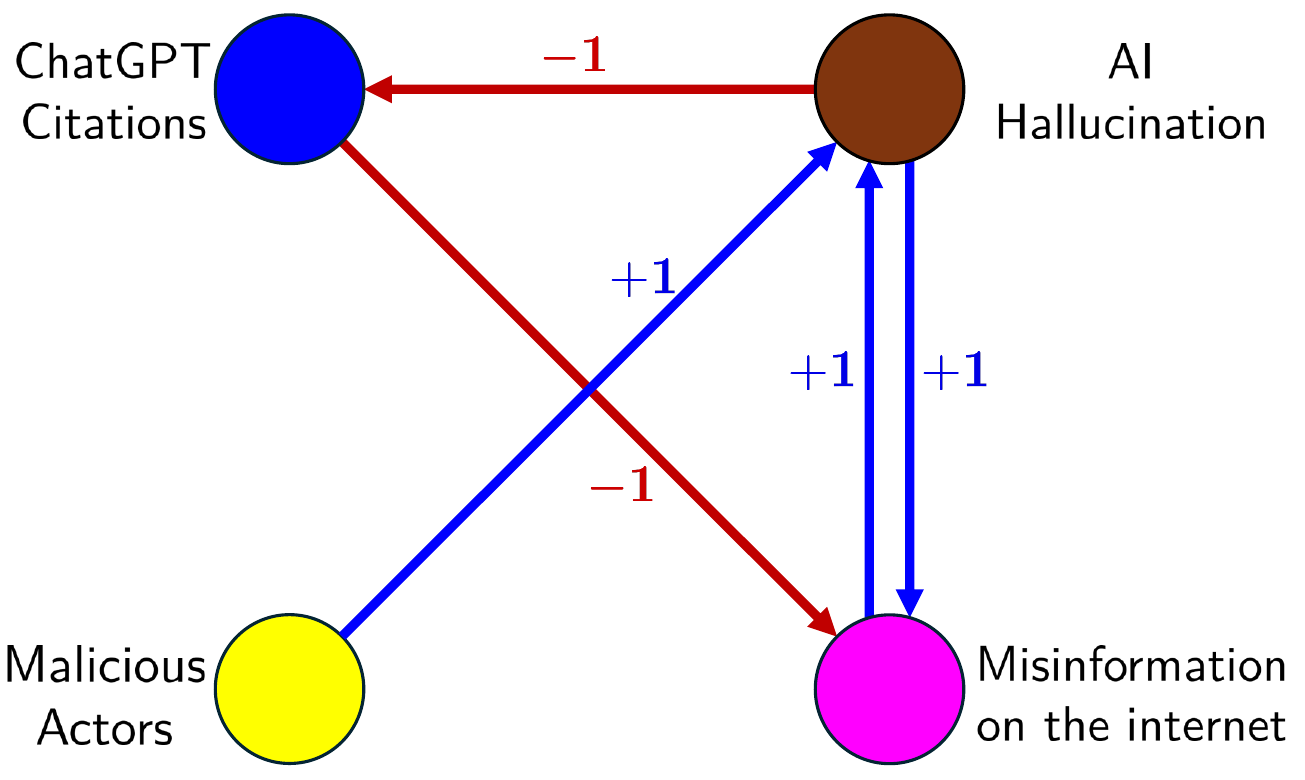}
\caption{\small{Human-generated FCM}}
\label{fig:Human_FCM}
\end{subfigure}
\begin{subfigure}{0.51\linewidth}
\vspace{0.2in}
\includegraphics[width=0.96\textwidth]{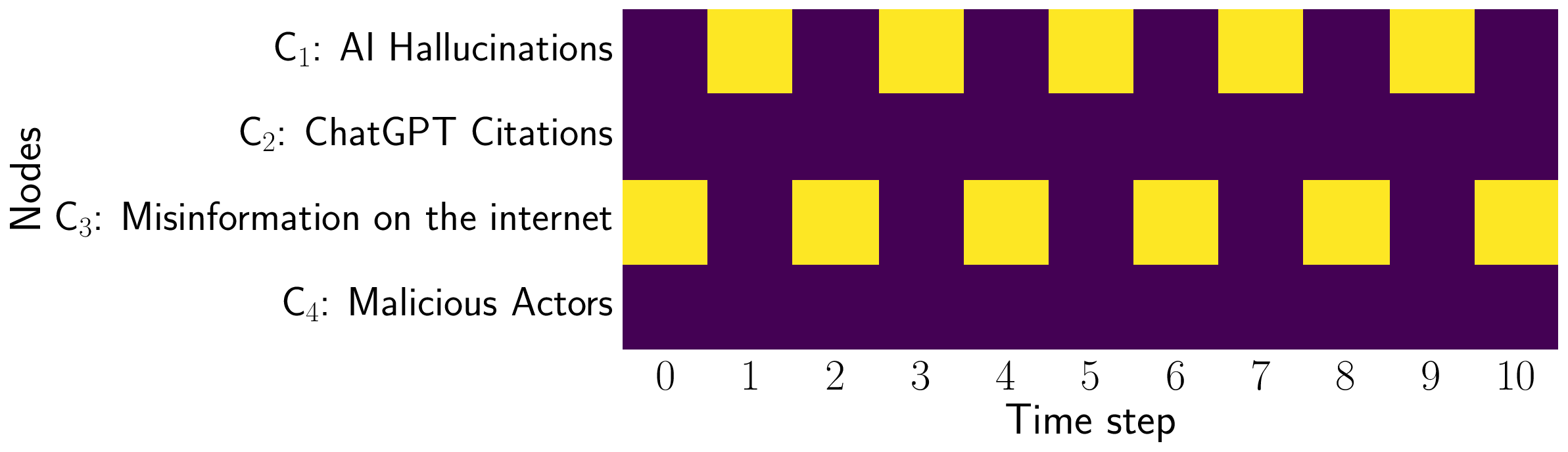} 
\caption{\small{Limit cycle: Human-generated FCM}}
\label{fig:Human_LimitCycles}
\end{subfigure}
\begin{subfigure}{0.5\linewidth}
\vspace{0.2in}
\includegraphics[width=0.96\textwidth]{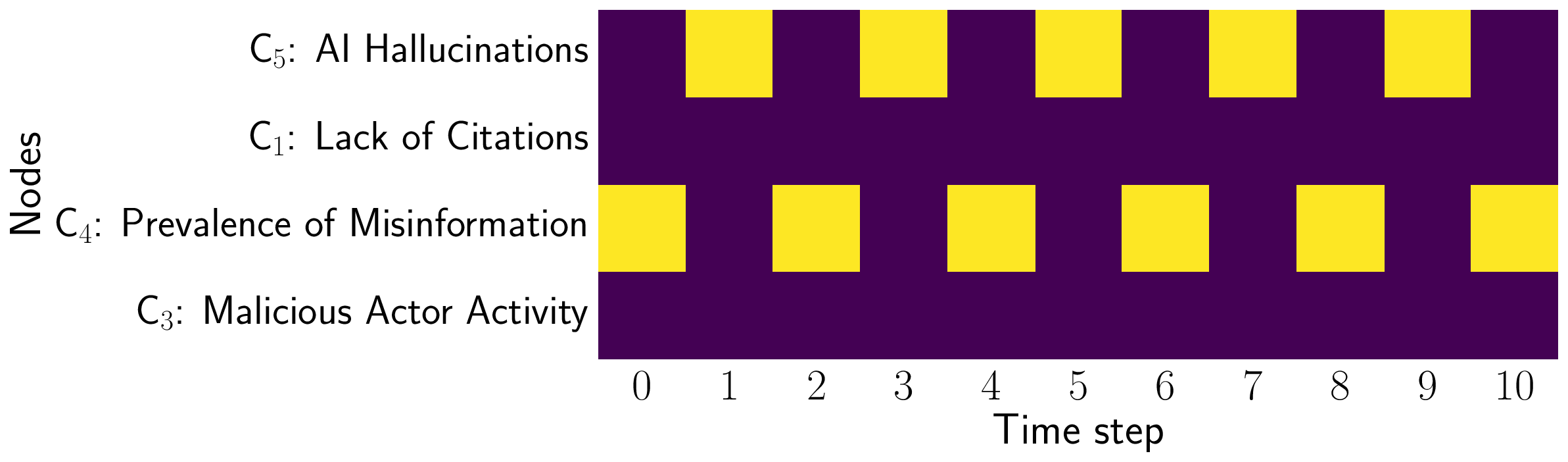} 
\caption{\small{Limit cycle: LLM-generated FCM}}
\label{fig:LLM_LimitCycles}
\end{subfigure}
\caption{\small{ FCM extraction from a Kissinger AI-hallucination text: (a) Human-generated FCM from the same ``AI hallucination'' input text. 
(b) Limit cycles from the human-generated FCM: 
The time steps are along the $x$-axis and the nodes are along the $y$-axis. 
Active nodes are in yellow and the inactive nodes are in purple. 
(c) Limit cycles from Gemini's extracted FCM for similar initial conditions. 
The figure only shows 4 nodes from Gemini's FCM that correspond to the nodes from human-generated FCM.}
  }
\label{fig:Human-FCM-v1}
\end{figure}

Section~\ref{simulations} shows examples of FCM extraction. 
Figures~\ref{fig:LLM-Agent-1}-\ref{fig:Gemini-Limit-Cycles} show this process for a one-paragraph input text. 
Figure~\ref{fig:Human-FCM-v1} compares the limit cycles of the extracted FCM to those of a human-generated FCM. 
The figure shows that both FCMs converge to the same limit cycle. 
Figure~\ref{fig:kissinger-FCM} shows the complete FCM extracted from Henry Kissinger's WSJ article. 
Figure~\ref{fig:kissinger-Limit-Cycle} shows the corresponding limit cycle that the author implied despite never explicitly mentioning it in the article.

Figure~\ref{fig:Gemini-GPT-FCM-Mixture} mixes the extracted-FCMs from the same article with the same guided prompts but used 2 different LLM agents: Gemini-2.5 Pro and ChatGPT-4.1. 
The 24-node mixed FCM used equal mixing weights for both FCMs. 
Figure~\ref{fig:Edge-matrix-mixing} mixes the corresponding edge matrices of the 2 mixture-component FCMs using zero-padding and convex combination. 
Figure~\ref{fig:fcm-mixture-limit-cycle} shows the limit cycles of this final mixed FCM.

\section{Using LLM Agents to Extract FCMs from Text}

This paper shows how to use LLM agents to extract FCM nodes and edges from raw text by using just the language of the text.
It does not rely on labels or annotations. 

The technique starts with the analysis of nouns and verbs in the target text.
The FCM nodes are usually nouns or noun phrases.
The noun terms can describe physical or social properties or policy variables or anything else that can exert or absorb causal effects. 
Verbs in the text describe FCM causal edges. 
The process of edge extraction is both transparent and interpretable because it gives direct evidence for the edges by quoting from the text itself. 

Our automated agentic approach differs from other efforts to extract FCMs from text that use LLMs in some way.  
Schuerkamp \cite{schuerkamp2025guiding} used LLMs and human input to extract causal edges from text using LLMs.
They used evolutionary algorithms to learn the edge weights while their list of nodes needed human input. 
Berijanian \cite{berijanian2024soft} also extracted FCMs from text sources using LLMs.
Their method relied on some initial human annotation of text with (“source”, “target”, “direction”) tuples.

Extracting FCM nodes from text somewhat resembles Named Entity Recognition tasks for LLMs\cite{wang2023gpt}. 
Here the LLMs identify key information from the text and classify them based on predefined categories. 
The LLMs can extract nouns and noun phrases from a text based on the properties of nouns in that language. 
But every noun or noun phrase in the text does not correspond to a node in the FCM. 
That approach would grow impossibly large and inaccurate FCMs.
Finely tuned system instructions can prompt the LLM to pick only the nouns that also share the properties of an FCM node. 

\begin{figure}[ht]
\centering
\includegraphics[width=0.8\textwidth]{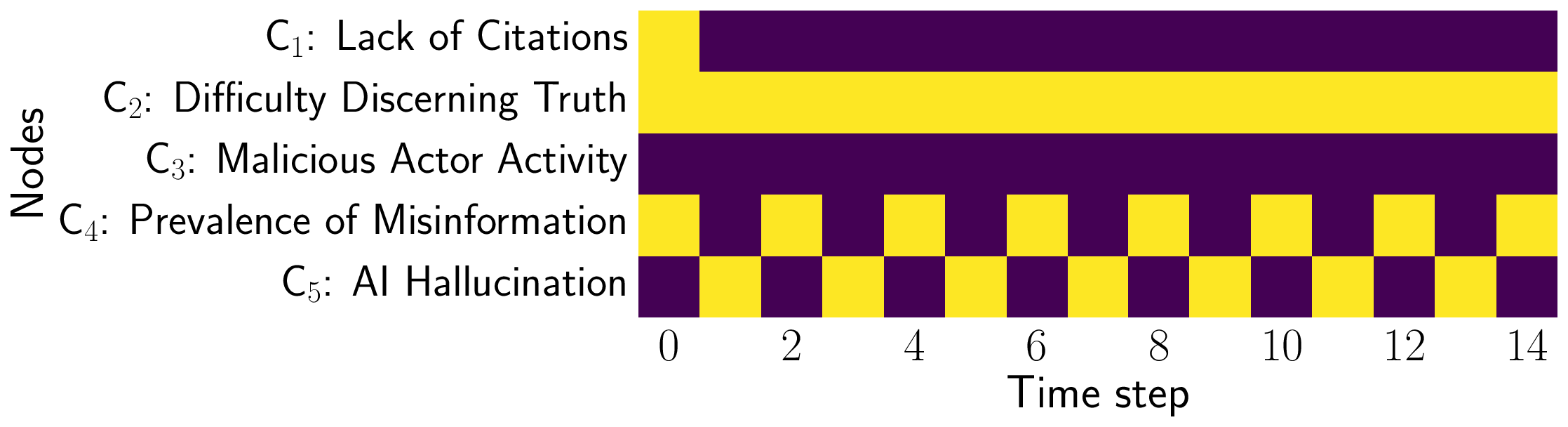}
\caption{The limit cycles of LLM's extracted FCM. The LLM's FCM converged to a 2-step limit cycle starting from the initial state $\begin{bmatrix} 1 & 1 & 0 & 1 & 0 \end{bmatrix}$. 
The nodes are along the $x$-axis and the time steps are along the $y$-axis. 
Active nodes are in yellow and the inactive nodes are in purple. }
\label{fig:Gemini-Limit-Cycles}
\end{figure}

Unguided prompts into an LLM are not reliable for extracting FCMs in general. 
They often have trouble distinguishing negative causal relationships from positive ones and might well inaccurately weight the FCM edges. 
They also often do not give reasons for the FCM edges. 
They may even hallucinate edges based on things not explicitly mentioned in the text itself. 

We design system instructions that force the LLMs to choose the nodes and edges in a systematic manner. 
This transparent 3-step process explains all the decisions made by the LLM and also reduces the chances and extent of hallucinations. 
The LLM supports all its decisions by quoting evidence from the text source. 

\section{FCM Basics}\label{FCM}

FCMs model causal dynamical systems as directed weighted graphs. 
The concept nodes describe the causal variables in the dynamical system and the directed edges describe the causal relationships between those nodes. 
FCMs allow feedback and therefore converge to non-trivial equilibria like limit cycles. 
FCMs model dynamical systems by approximating their underlying maps from inputs to equilibria. 

\begin{figure}[ht]

\end{figure}

\subsection{Causal Edge Matrix $E$}
The directed edges of the FCM describe the causal relationships between concept nodes. 
An edge $e_{ij}$ from the $i^{\text{th}}$ concept node $C_i$ to the $j^{\text{th}}$ concept node $C_j$ says ``$C_i$ causes $C_j$''. 
The fuzzy edge-weights describe partial causality. 
The edge weight $w_{ij} \in [-1,1]$ on the edge $e_{ij}$ gives the degree to which $C_i$ causes $C_j$:
\begin{align}
    w_{ij} = degree(C_i \rightarrow C_j).
\end{align}
A positive $w_{ij}$ says that $C_j$ increases when $C_i$ increases. 
A negative $w_{ij}$ means that $C_j$ decreases if $C_i$ increases. 
The weight $w_{ij}$ is zero when there is no causal edge between $C_i$ and $C_j$. 
The magnitude of $w_{ij}$ is high when there is a strong causal relationship between $C_i$ and $C_j$. 
Low magnitude of $w_{ij}$ describes a weak causal relationship between $C_i$ and $C_j$.

An $n \times n$ matrix $E$ describes all the directed weighted edges of a $n$-node FCM. 
The edge weight $w_{ij}$ corresponds to the matrix element on the $i^{\text{th}}$ row and the $j^{\text{th}}$ column. 
The matrix element is zero if there is no edge between the corresponding node-pair.

\subsection{Discrete-Time Evolution of FCMs}

A $n$-dimensional row vector $C(t) \in [0,1]^n$ describes the state of the FCM's concept nodes at time $t$. 
The $i^{\text{th}}$ node is ``active'' at time $t$ if the $i^{\text{th}}$ component $C_i(t)$ of the state vector $C(t)$ is equal to or close to one. 
The $i^{\text{th}}$ node is ``inactive'' at time $t$ if $C_i(t)$ is equal to or close to zero. 
A node is partially active otherwise. 
The causal variables corresponding to the active nodes are present in the system and those corresponding to the inactive nodes are absent. 
The causal factors are partially present in the system if their corresponding nodes are partially active. 

FCMs evolve in discrete time through vector-matrix multiplication and nonlinear squashing. 
The state $C_j(t+1)$ of the $j^{\text{th}}$ concept node $C_j$ at discrete time step $t+1$ is:
\begin{align}\label{eq:fcm-update}
    C_j(t+1) = \Phi\bigg(\sum_{i=1}^n  C_i(t)w_{ij}\bigg)
\end{align}
where $\Phi$ is a nonlinear function bounded between zero and one. 

The sum $\sum_{i=1}^n  C_i(t) w_{ij}$ is the matrix product between the state row-vector $C(t)$ and the edge matrix $E$. 
The nonlinear function $\Phi$ then squashes this product between zero and one.


\begin{figure}[ht]
\centering
\includegraphics[width=0.95\textwidth]{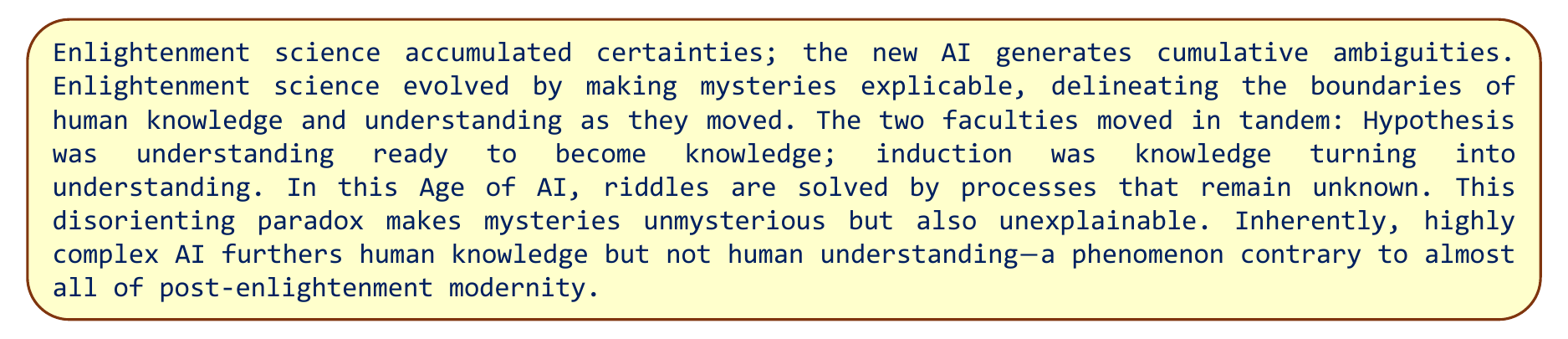}
\caption{A paragraph from the WSJ article \emph{ChatGPT Heralds an Intellectual Revolution} by Henry Kissinger et al. }
\label{fig:Kissinger-Text}
\end{figure}

This process repeats itself to give the discrete-time evolution of the FCM. 
The FCM starts with the initial state $C(0)$ at time $t = 0$ and then goes through the states $C(1)$, $C(2)$, $C(3)$, and so on in order. 
The active nodes in this state-vector sequence qualitatively describe the trajectory of the dynamical system that the FCM models.

\begin{figure}[!ht]
\begin{subfigure}{0.4\linewidth}
\centering
\includegraphics[width=0.96\textwidth, height=0.96\textwidth]{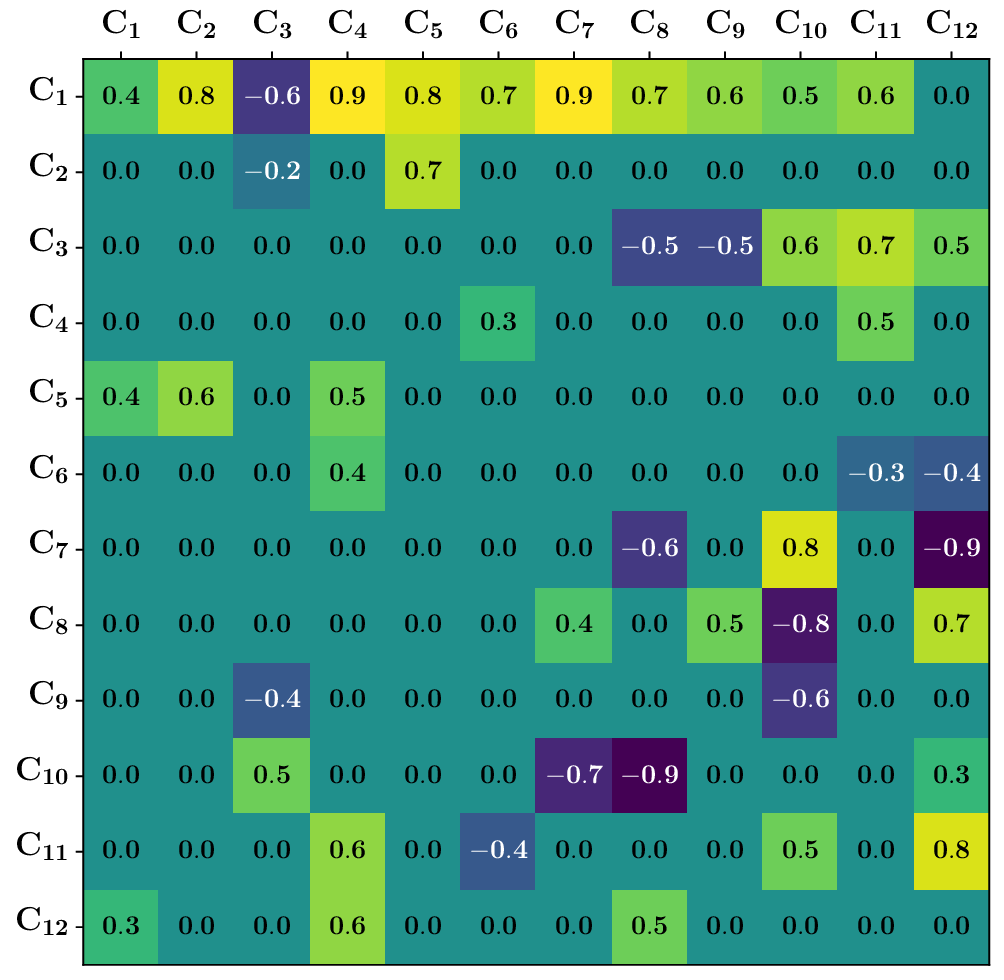}
\caption{\small{Unguided prompt}}
\label{fig:Gemini_baseline}
\end{subfigure}
\hfill
\begin{subfigure}{0.4\linewidth}
\centering
\includegraphics[width=0.96\textwidth, height=0.96\textwidth]{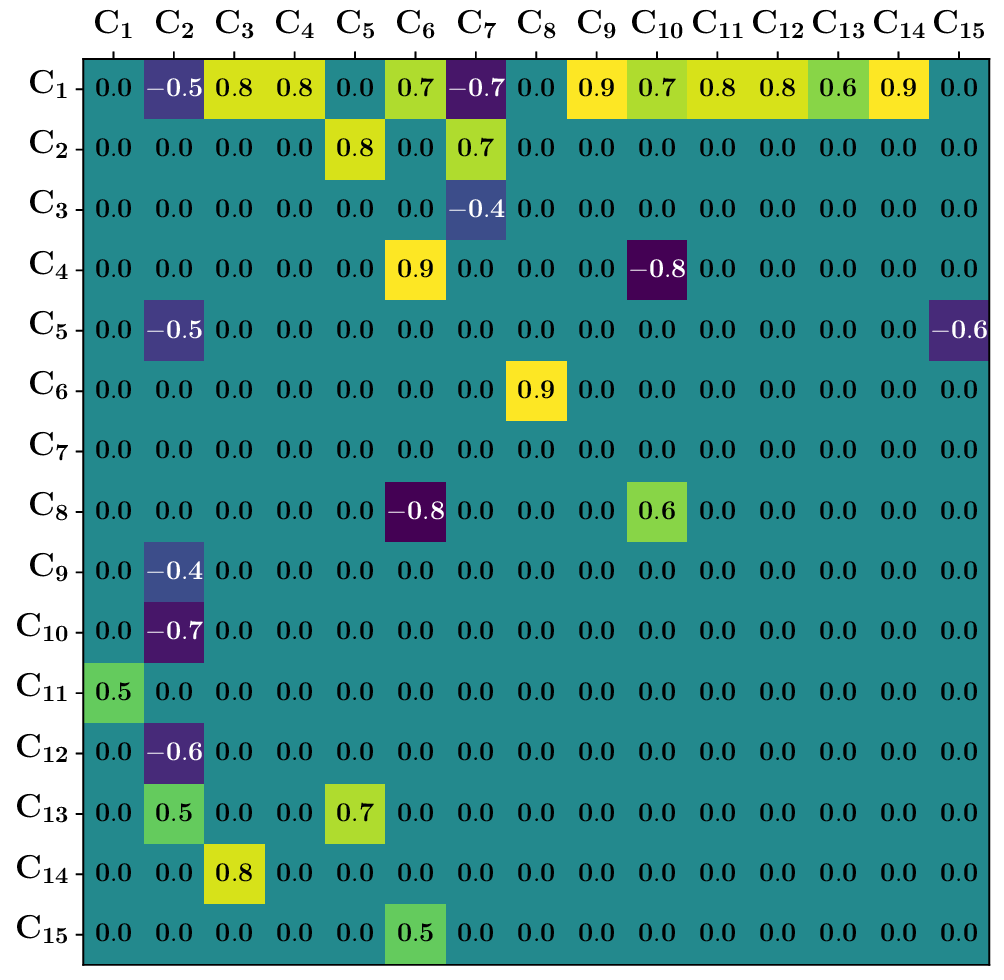} 
\caption{\small{Guided prompt}}
\label{fig:Gemini_custom}
\end{subfigure}
\caption{\small FCM with Gemini-2.5 Pro: 
(a) The edge matrix corresponding to the 12-node Kissinger AI FCM from an unguided prompt. (b) The edge matrix corresponding to the 15-node Kissinger AI FCM from a guided prompt. 
  }
\label{fig:gemini_fcm}
\end{figure}

\begin{figure}[ht]
\begin{subfigure}{0.48\linewidth}
\centering
\includegraphics[width=0.96\textwidth, height=0.96\textwidth]{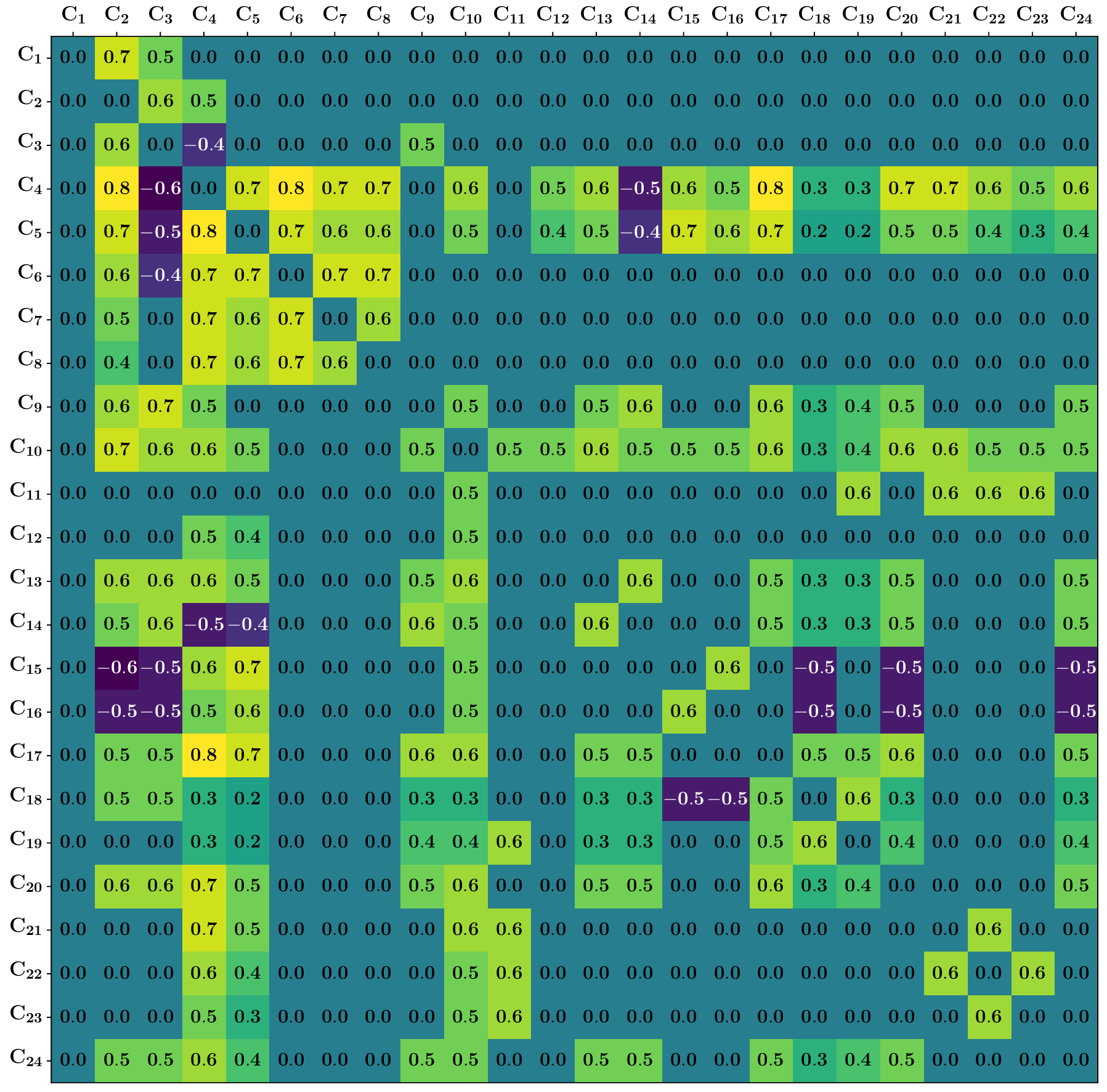}
\caption{\small{Unguided prompt}}
\label{fig:GPT_baseline}
\end{subfigure}
\begin{subfigure}{0.48\linewidth}
\centering
\includegraphics[width=0.96\textwidth, height=0.96\textwidth]{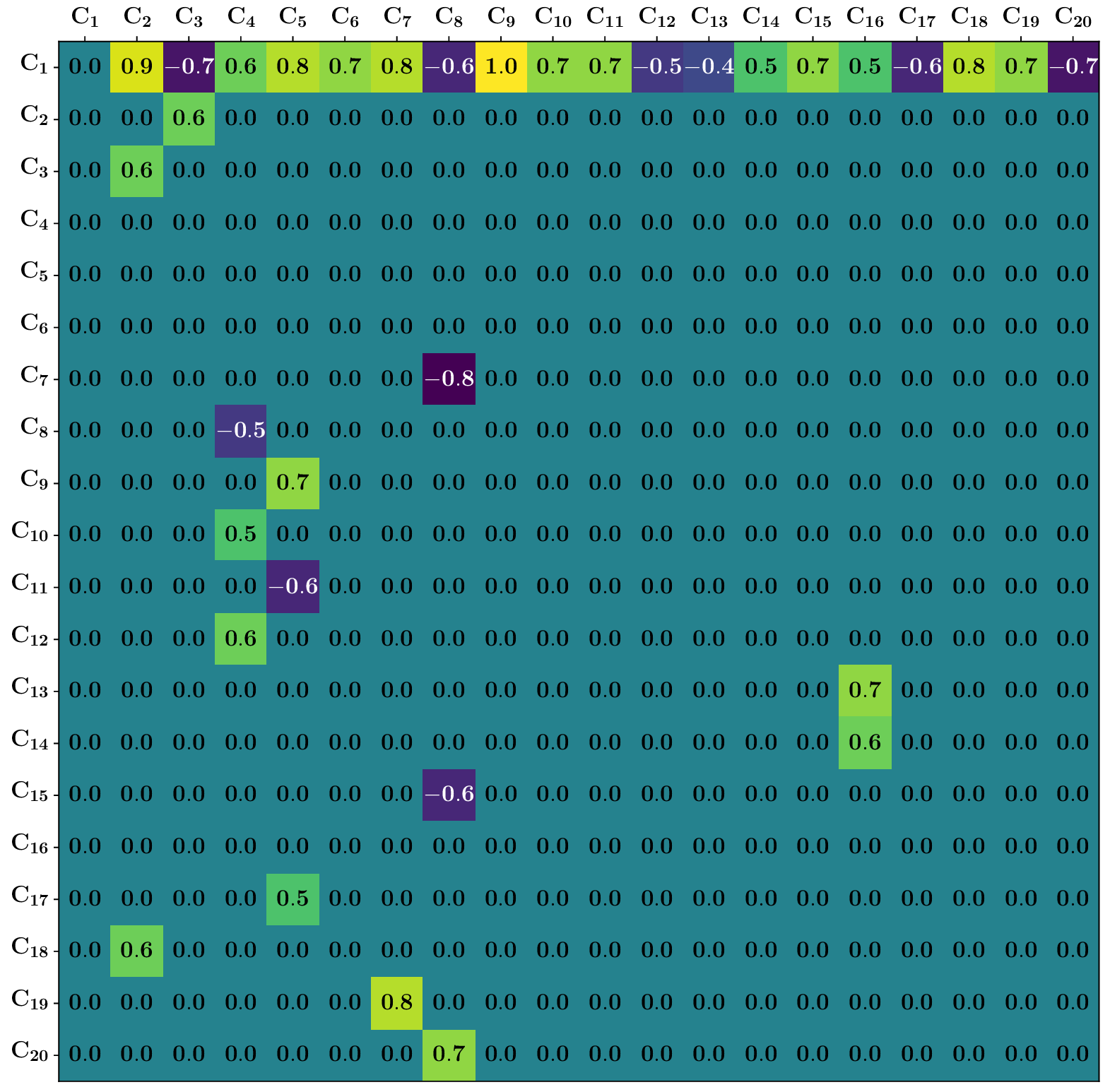} 
\caption{\small{Guided prompt}}
\label{fig:GPT_custom}
\end{subfigure}
\caption{\small FCM with GPT-4.1: 
(a) The edge matrix corresponding to the 24-node Kissinger AI FCM with an unguided prompt . (b) The edge matrix corresponding to the 20-node Kissinger AI FCM with a guided prompt. 
  }
\label{fig:GPT_fcm}
\end{figure}

\subsection{FCM Equilibria}

The equilibria characterize a dynamical system. 
The equilibrium behavior of the FCM depends on the limiting behavior of the state-vector sequence. 
The FCM converges to a ``fixed point'' if the state-vector sequence converges to a constant vector. 
The FCM converges to a $K$-step ``limit cycle'' for an integer $K > 1$ if $C(t+K) = C(t)$ somewhere in the state-vector sequence. 
Then the FCM converges to an equilibrium where $K$ state vectors repeat themselves over and over in the same order. 
The FCM may also converge to a chaotic attractor where there are no repeating patterns in the state-vector sequence. 

The set of all initial conditions $C(0)$ that lead to a given equilibrium describes the ``basin of attraction'' for that equilibrium. 
The FCM describes a map from these basins to their corresponding equilibrium attractors. 
The basins of the FCM's equilibria partition the FCM's input space. 
The FCM models a dynamical system by approximating its corresponding basin-to-equilibrium map. 

\subsection{FCM Mixtures}

FCMs combine their knowledge through convex mixing. 
Consider $m$ FCMs. 
Say the $k^{\text{th}}$ FCM has the set of nodes $S_k$ and the edge matrix $E_k$. 
The node-set $S$ for the $N$-node FCM mixture is $S_1\cup S_2\cup S_3\cup ...\cup S_m$. 
The $N\times N$ matrix $\Tilde{E_k}$ pads the $k^{\text{th}}$ edge matrix $E_k$ with zero rows and zero columns corresponding to the nodes in the set difference $S - S_k$. 
The edge matrix $E$ for the mixed FCM is
\begin{align}
    E = \sum_{k=1}^m v_k\Tilde{E_k} \label{eg:fcm_mixing}
\end{align}
where $v_k$ are convex mixing weights such that $v_k\geq0$ and $\sum_{k=1}^mv_k=1$. 
FCM mixing is closed.
This means that mixing FCMs gives back an FCM. 
Figure~\ref{fig:Edge-matrix-mixing} shows this process for a 2-component equal-weighted FCM-mixture.

\section{LLM-Based FCM Extractor}\label{LLM-FCM-Extractor}

Text articles often describe causal dynamical systems without explicit mathematical formulation. 
This makes it hard to understand what dynamics the underlying dynamical system predicts. 
It is also difficult to verify the author's guess regarding the dynamical system's behavior. 
A human expert can read through the text and model the dynamical system it describes. 
But this may not always be possible, may take too long, or may be too expensive. 
This motivates automated dynamical-system modeling from text. 

LLMs are good at parsing text and processing natural language. 
They can extract causal information from text that describes dynamical systems and use it to model the underlying dynamical systems as FCMs. 
But LLMs hallucinate with unguided prompts. 
Their answer may also vary across multiple runs with the same prompt. 
The LLMs need guiding prompts that are designed to extract FCMs.

System instructions manipulate an LLM agent to behave a certain way. 
System instructions tell the LLM how to process the input and how to structure the output. 
The same LLM can react differently to ``prompts'' based on different system instructions. 
A sequence of appropriate system instructions can guide an LLM agent to systematically extract FCM-nodes and edges from text input. 

\begin{figure}[ht]
\centering
\includegraphics[width=0.9\textwidth]{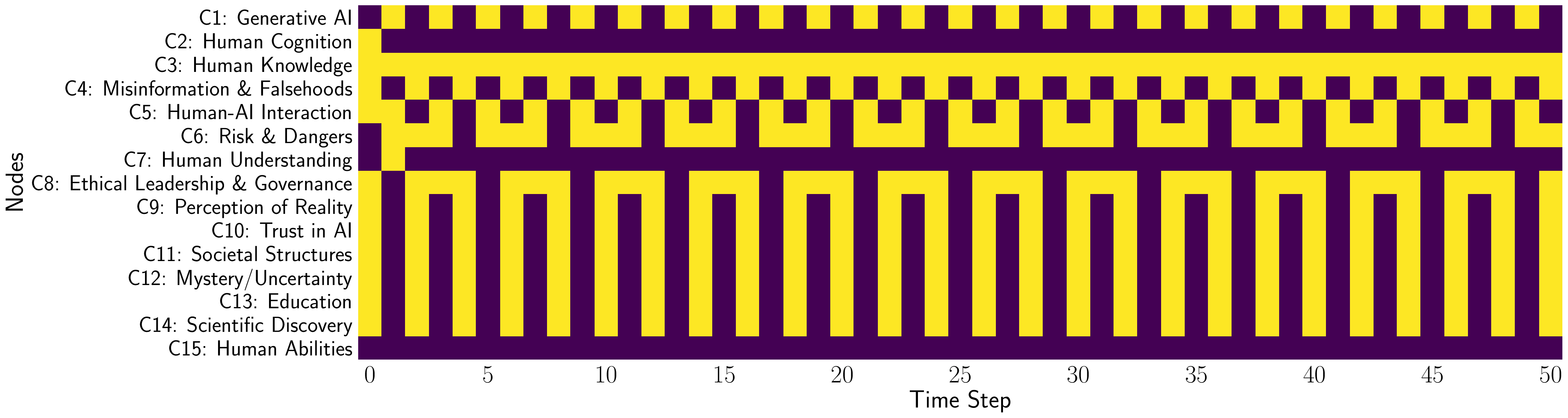}
\caption{Limit cycles from the extracted FCM. 
The time steps are along the $x$-axis and the nodes are along the $y$-axis. 
Active nodes are in yellow and the inactive nodes are in purple. }
\label{fig:kissinger-Limit-Cycle}
\end{figure}

\subsection{Step 1: Noun Extraction}

The nodes of an FCM are present in the text as nouns and noun phrases. 
System instructions ask the LLM Agent to go through the text sentence-by-sentence and make a list of all nouns, noun phrases and pronouns present in the text. The agent then replaces all the pronouns with their corresponding antecedents as well.

\begin{figure}[ht]
\centering
\includegraphics[width=0.8\textwidth]{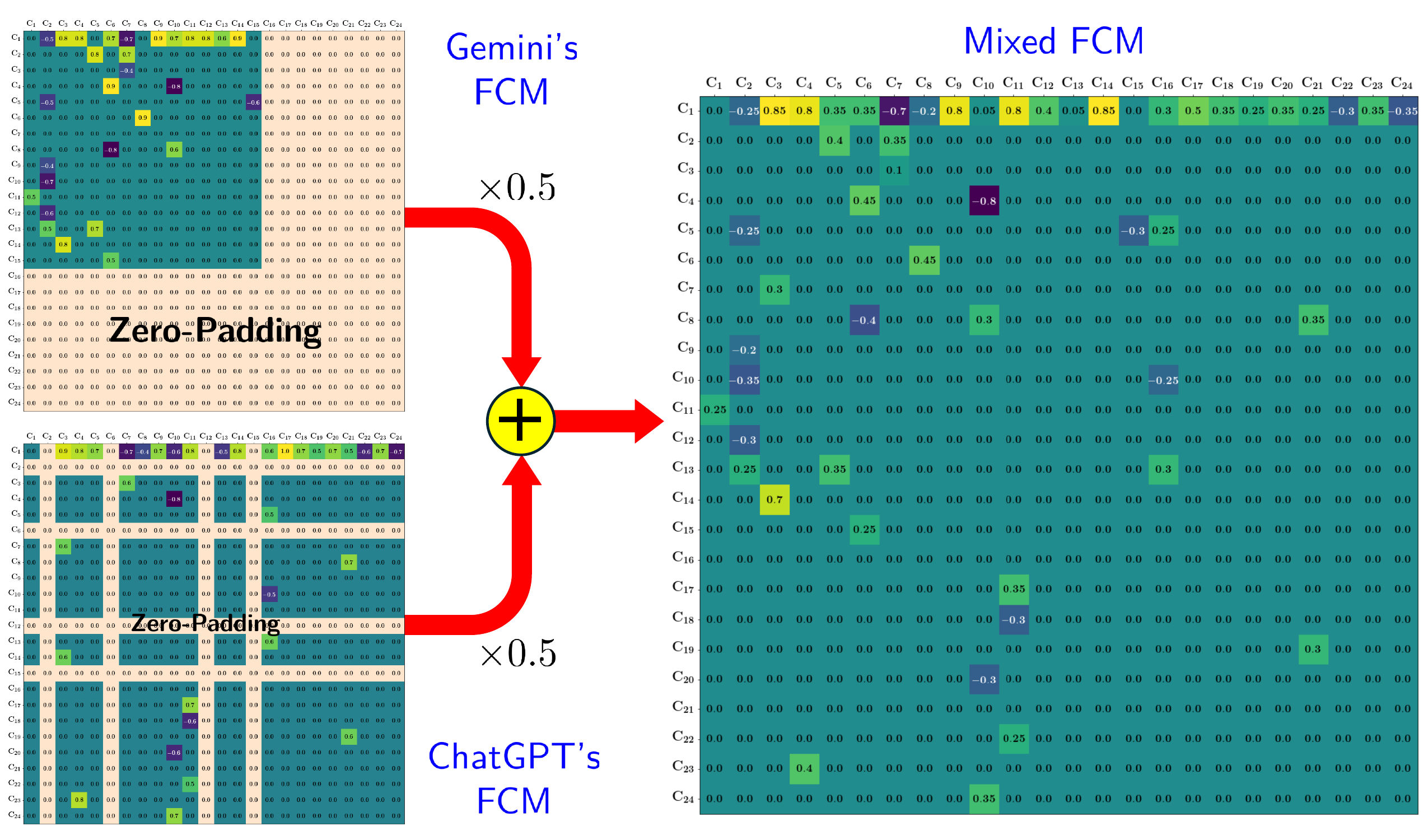}
\caption{FCMs' edge matrices mix through zero-padding and convex combination described in equation (\ref{eg:fcm_mixing}). 
The edge matrix of the 15-node Gemini-extracted FCM is on the top-left and the edge matrix of the 20-node ChatGPT-extracted FCM is on the bottom-left. 
The edge matrix of the 24-node equal-weight FCM-mixture is on the right. 
The bigger edge weights are colored brighter in the matrices. 
The matrices on the left also highlight the zero-padded rows and columns of the mixture components. 
}
\label{fig:Edge-matrix-mixing}
\end{figure}

\begin{figure}[ht]
\centering
\includegraphics[width=0.8\textwidth]{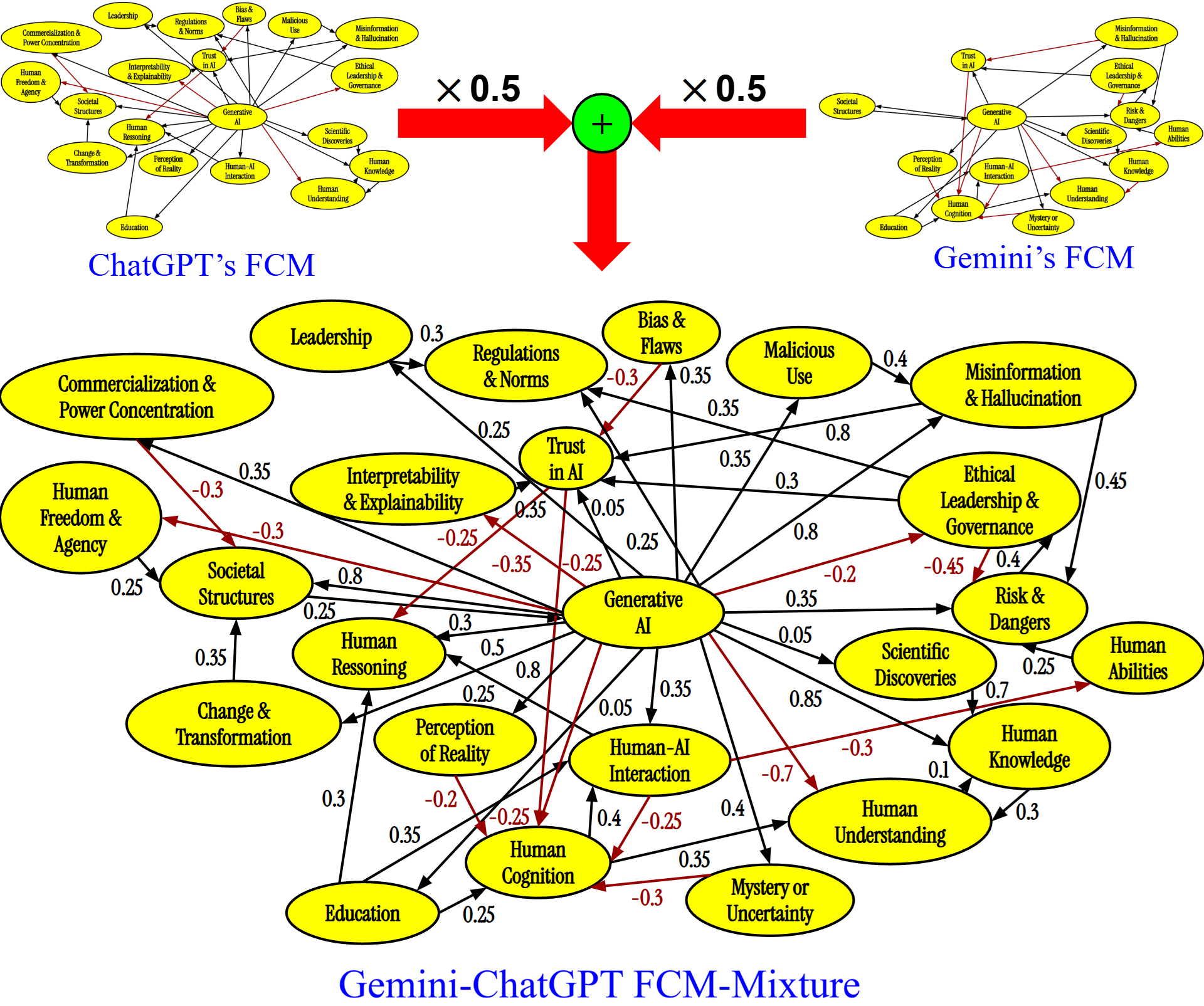}
\caption{Mixed FCMs:  An equal-weight mixture of the 15-node Gemini-extracted FCM and the 20-node ChatGPT-extracted FCM after appropriate zero-padding of rows and columns. 
The final mixed FCM has 24 nodes and 52 edges.
The 15-node Gemini-extracted FCM is on the top-right and the 20-node ChatGPT-extracted FCM is on the top-left.
}
\label{fig:Gemini-GPT-FCM-Mixture}
\end{figure}

\subsection{Step 2: Noun Refinement}

FCM nodes are associated with a qualitative or quantitative measure. 
This measure can causally increase or decrease due to other nodes. 
The $2^{\text{nd}}$ system instruction to the LLM agent asks it to filter the list of nouns and noun phrases and only keep nouns that are associated with qualitative or quantitative measures and are causally connected to other nouns with similar measures. 
It then returns a refined list of nouns and noun phrases that will serve as the nodes of the FCM. 

\subsection{Step 3: Edge Extraction}

The LLM agent with the $3^{\text{rd}}$ set of system instructions pairs every node from the refined list with every other node. 
It then goes through every node-pair and looks for verbs from the text to establish a positive, negative, or zero causal connection. 
The agent also gives reason for each edge by directly quoting from the text. 
It also assigns a weight to the edge based on the verb used in the text. 
This gives us the complete FCM described by the text. 

Equation~(3) then gives the limit cycles of the FCM. 
These are the dynamical equilibria that the text suggests.

\section{Simulations}\label{simulations}

\begin{figure}[ht]
\centering
\includegraphics[width=\textwidth]{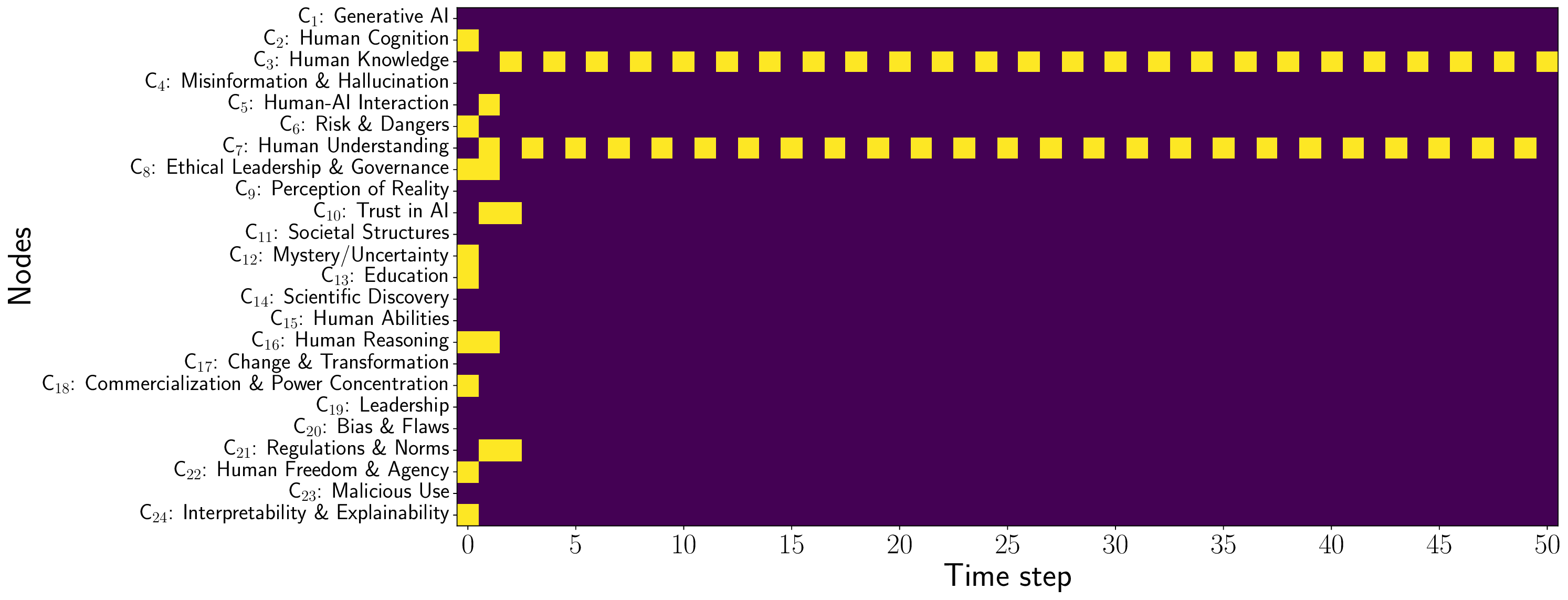}
\caption{Limit cycle from the mixed Gemini-ChatGPT FCM. 
The time steps are along the $x$-axis and the nodes are along the $y$-axis. 
Active nodes are in yellow and the inactive nodes are in purple. 
The 2-step limit cycle oscillates between states ``Human Knowledge'' and ``Human Understanding.'' }
\label{fig:fcm-mixture-limit-cycle}
\end{figure}

 Gemini-2.5 Pro  LLM with ``temperature = 1'' and ``top $p$ = 0.95'' took a paragraph of text from the Kissinger article describing the effects of misinformation on the internet on AI hallucinations as input. 
It then followed system instructions 1-3 respectively to systematically extract the nouns, nodes, and causal edges. 
This gave a FCM with 5 nodes and 6 edges. 

The LLM highlighted the nouns and noun phrases directly from the text as reasons for choosing certain nodes. 
Figures~\ref{fig:LLM-Agent-1} and \ref{fig:LLM-Agent-2} highlight these nouns and noun phrases in red. 
The node ``{\emph{Malicious Actor Activity}}'' for example comes from the noun phrase ``\emph{malicious actors}''. 
The negative edge from ``\emph{lack of citations}'' to ``\emph{Difficulty discerning truth}'' comes from the sentence ``\emph{The lack of citations in ChatGPT's answers makes it difficult to discern truth from misinformation}''. 
Figures~\ref{fig:LLM-Agent-1}-\ref{fig:LLM-Agent-3} show this 3-step process step-by-step.

This FCM converged to fixed points or limit cycles depending on the initial state. 
AI hallucinations spread misinformation over the internet. 
AI systems then train on this misinformation which causes them to hallucinate. 
Both steps make truth harder to discern.  
Figure~\ref{fig:Gemini-Limit-Cycles} also shows a 2-step limit cycle of the extracted FCM.

A human took the same text as input but came up with an FCM with only 4 nodes and 5 edges. 
The 4 human-generated nodes ``{\emph{AI Hallucinations}}'', ``{\emph{Misinformation on the internet}}'', ``{\emph{ChatGPT Citations}}'', and ``{\emph{Malicious Actors}}'' are also present in the LLM-extracted FCM. 
The node ``{\emph{ChatGPT Citations}}'' plays the opposite role of ``{\emph{Lack of Citations}}''. 
This human-generated FCM also converged to a 2-step limit cycle. 
Figure~\ref{fig:Human-FCM-v1} compares this limit cycle to that of the LLM-extracted FCM by only considering the nodes that are present in both the FCMs. 
It shows that both FCMs converge to the same limit cycle. 

The same LLM then took a WSJ article \emph{``ChatGPT Heralds an Intellectual Revolution''} by Henry Kissinger et al. as input. 
Figure~\ref{fig:Kissinger-Text} shows one paragraph from this article.
An unguided-prompt to the LLM extracted a FCM with 12 nodes. Figure~\ref{fig:Gemini_baseline} shows the edge matrix corresponding to this FCM. 
A guided prompt to the same LLM gave a 15-node FCM instead.
Figure~\ref{fig:kissinger-FCM} shows this FCM and Figure~\ref{fig:Gemini_custom} gives its edge matrix. 
This 15-node FCM converged to a 4-step limit cycle. 
Figure~\ref{fig:kissinger-Limit-Cycle} shows this limit cycle. 

This limit cycle also tells a what-if story of the underlying dynamical system. 
Generative AI gets widely used, human-AI interactions rise, and human knowledge grows. 
This helps leaders govern ethically but it also comes with risks and dangers.
People trust the mysterious and uncertain AI, misinformation and falsehoods spread, and society changes. 
But AI improves education at the same time leading to scientific discoveries. 
Generative AI is used widely again but this time without ethical leadership. 
Society changes again but this time without generative AI. 
This limit cycle was not mentioned in the source article at all but the article did imply it through its discussion on the causal variables and their relationships.

OpenAI's ChatGPT-4.1 LLM agent respectively produced a 24-node FCM and a 20-node FCM based on the same unguided and guided prompts. 
Figure~\ref{fig:GPT_fcm} shows the edge matrices corresponding to these FCMs.
The 15-node Gemini-FCM and the 20-node ChatGPT-FCM mixed with equal mixing weights to give the final 24-node mixed FCM. 
Figure~\ref{fig:Gemini-GPT-FCM-Mixture} shows this FCM-mixture. 
Figure~\ref{fig:Edge-matrix-mixing} shows its edge matrix while Figure~\ref{fig:fcm-mixture-limit-cycle} shows its limit cycle.

\section{Conclusions}

This paper shows how iterated and structured LLM agents can generate feedback causal fuzzy cognitive maps from text documents or from transcribed speech.  Further causal structuring should produce larger FCMs and more powerful LLMs should produce richer and more accurate FCMs.  
The process scales across text sources and knowledge bases because causal language is fairly standard and because FCM operation depends only vector-matrix multiplication and simple nonlinear units.  

The mixing structure of FCMs also allows a given large text to break into chunks and subchunks with corresponding mixed FCM components.
This process can help with node and edge identification in large documents.
Mixing such FCMs still produces an FCM and this should further help the growth and uses of very-large-scale causal knowledge networks.
More advanced agentic FCM systems will lengthen the agentic leash.
This will allow the knowledge system to grow and use complex causal networks as the system both shapes and obeys the evolving global equilibria of the total FCM dynamical system.
%
%
%
%
\bibliographystyle{splncs04}
\bibliography{bibdata}
\end{document}